\pdfoutput=1

\documentclass[11pt]{article}

\usepackage[final]{acl}

\usepackage{times}
\usepackage{latexsym}

\usepackage[T1]{fontenc}

\usepackage[utf8]{inputenc}

\usepackage{microtype}

\usepackage{inconsolata}

\usepackage{graphicx}

\usepackage{times}
\usepackage{latexsym}
\usepackage{amsmath}

\usepackage[T1]{fontenc}

\usepackage[utf8]{inputenc}

\usepackage{microtype}

\usepackage{inconsolata}

\usepackage{graphicx}
\usepackage{booktabs}
\usepackage{enumitem}

\title{The State and Fate of Summarization Datasets: A Survey}

\author{Noam Dahan \and  Gabriel Stanovsky \\
       The Hebrew University of Jerusalem \\ \href{mailto:noam.dahan1@mail.huji.ac.il}{\texttt{noam.dahan1@mail.huji.ac.il}}}

\begin{document}
\maketitle
\begin{abstract}
Automatic summarization has consistently attracted attention due to its versatility and wide application in various downstream tasks. Despite its popularity, we find that annotation efforts have largely been disjointed, and have lacked common terminology. Consequently, it is challenging to discover existing resources or identify coherent research directions. To address this, we survey a large body of work spanning 133 datasets in over 100 languages, creating a novel ontology covering sample properties, collection methods and distribution. 
With this ontology we make key observations, including the lack of accessible high-quality datasets for low-resource languages, and the field's over-reliance on the news domain and on automatically collected distant supervision. Finally, we make available a web interface that allows users to interact and explore our ontology and dataset collection,\footnote{\url{https://github.com/edahanoam/Awesome-Summarization-Datasets}}\textsuperscript{,}\footnote{\url{https://searchdatasets-fru5zwwfm2shmyrtna9gjh.streamlit.app/}} as well as a template for a \emph{summarization data card}, which can be used to streamline future research into a more coherent body of work.
\end{abstract}

\section{Introduction}
\begin{figure}
\centering
    \includegraphics[width=0.9\columnwidth]{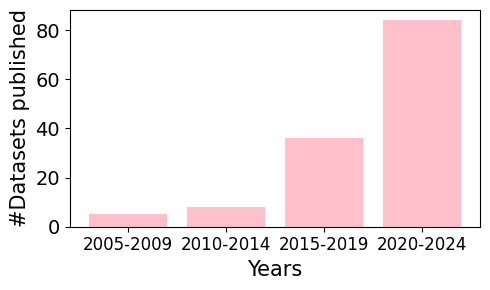}

    \caption{The number of summarization datasets being published is continuously increasing, yet the field lacks standardization and common terminology.
    } 
    \label{fig:dataset_by_year}
\end{figure}

Summarization is the task of shortening a text while preserving the most important information it contains. This task has been explored for over 60 years~\cite{luhn1958automatic} yet it remains a longstanding challenge, as evidenced by the steadily increasing number of summarization datasets released every year (see Figure~\ref{fig:dataset_by_year}). 

Despite the many research efforts in creating summarization datasets, we find they lack standardization and common terminology. This shortcoming makes it difficult to discover existing resources and bottlenecks, and to identify general directions for future research.

In this work, we conduct the most comprehensive survey of summarization datasets to date, covering a wide collection of 133 datasets across 104 languages. In Section~\ref{Ontology}, we develop a comprehensive ontology geared towards modern summarization datasets, positioning each dataset along seven different scales, including information about the samples (language, language modality, domain and summarization shape) as well as collection methods (annotation efforts and sources of supervision), and the manner in which the dataset is distributed. 



In Section~\ref{Findings} we show that our proposed ontology surfaces five major trends and gaps in the field. First, we find that the lack of common terminology thus far often leads to inconsistency and to the omission of important details when presenting datasets, making comparison challenging. An important example of this inconsistency is the traditional abstractive-extractive distinction, which we argue has gradually shifted away from a binary trait to a more nuanced spectrum.

Second, we observe that the field is over-reliant on the news domain, which acts as a double-edged sword -- being easy to obtain while prone to low quality. This is especially evident for low-resource languages, where the vast majority of datasets are based on news articles.

Third, we find that the datasets available for low-resource languages tend to be of subpar standard and especially challenging to discover. This is because many languages are currently supported only through multilingual datasets, which tend to compromise quality for diversity~\cite{urlana2022tesum}, while also making it harder to know which languages are annotated in which dataset.

Fourth, we find that while the majority of annotations are gathered automatically, they are typically not originally created to serve as summaries, relying instead on distant supervision.

Fifth, we observe that copyright concerns, especially in the news domain, have been changing data distribution, with many datasets moving from publishing the data to publishing scripts that reconstruct it from URLs. 
Furthermore, a few seminal summarization datasets that were published publicly have been either taken down or are considered legally questionable due to copyright violations.

In Section~\ref{sec:fate} we publish two resources to standardize and help organize the vast research in the field. First, we make our dataset collection publicly available through a web UI which allows users to search datasets matching different criteria according to our terminology, for example, retrieving all manually created datasets in a certain language. This UI can help researchers discover datasets, as well as identify new research directions. We welcome community contributions to keep this resource updated with new research.

Second, inspired by  work on model cards~\cite{model-cards}, we propose a standard \textit{summarization data card} which covers the different items in our ontology as well as additional important properties. This can benefit future dataset developers in making their work more accessible, easy to discover, and comparable against other datasets.

We hope that our work helps chart the way for future work in the field, as well as enable researchers from various disciplines in exploring the wide range of existing summarization datasets.

To conclude we make the following contributions:

\begin{itemize}
    \item We formulate an ontology for summarization datasets, covering various axes regarding the samples, their collection and distribution.
    \item We survey a wide range of 133 datasets covering 104 languages, showing that our ontology is rich enough to surface five important trends and challenges, including the lack of terminology, the subpar quality of datasets in low-resource languages, and more.
    \item We make available two resources -- a web interface to interact with our survey and discover summarization datasets, and a formal \emph{summarization data card}, which will help streamline future datasets.
\end{itemize}

\section{Dataset Collection}
\label{Collection}






Our survey covers 133 datasets collected over nearly 20 years, spanning from 2005 to 2024. The full list of references can be found in Appendix \ref{sec:appendix_langs}.

\paragraph{Venues.}
Most datasets are drawn from papers published in top-tier NLP, ML and AI venues and workshops.\footnote{Venues: ACL, NAACL, EMNLP, LREC, EACL, SIGIR, AAAI, COLING, IEEE, NeurIPS, CoNLL, and TACL.} We also include datasets published through shared tasks and in language-specific journals and workshops. 

\paragraph{Filtering.}
We manually verify that each dataset is available for use, and exclude papers describing currently inaccessible resources. We include papers that explicitly focused on creating summarization datasets or that focused on summarization models and produced data as a by-product.

\paragraph{Collected Data.}
For each dataset, we gather information on its properties, collection methods and distribution. This methodology allows categorizing the current resources available, analyzing emerging trends and identifying the gaps in the field.


\section{Ontology of Summarization Datasets}
\label{Ontology}

\begin{figure*}[tb!]
    \centering 
    \includegraphics[width=\textwidth]{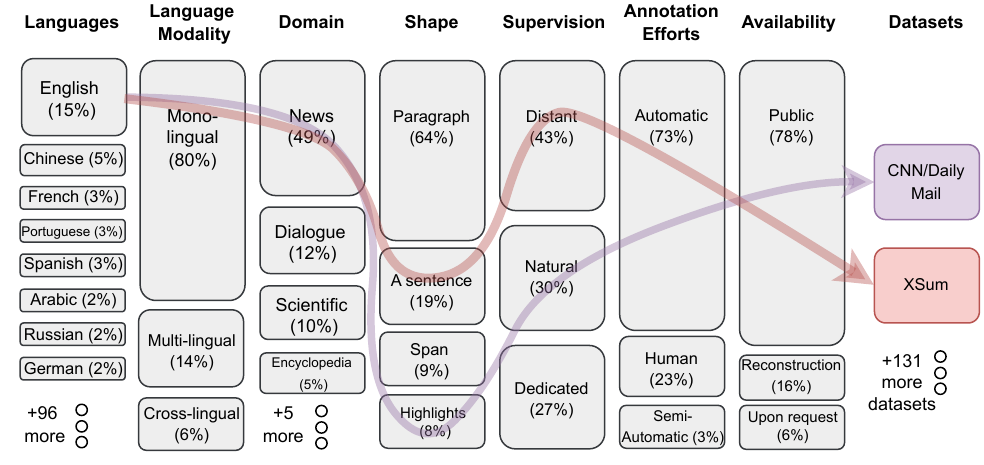}  
    \caption{Our ontology for summarization datasets, accompanied by our annotations. The percentages in the language column indicate the proportion of datasets that support each language, where we count each multilingual dataset as multiple monolingual datasets. The arrows showcase the decision pathways for selecting specific datasets. 
    }
    \label{fig:onto}

\end{figure*}

We propose a categorization of summarization datasets across seven orthogonal axes, and provide an ontology for each, depicted in Figure~\ref{fig:onto}, and elaborated below. This ontology helps contextualize the extensive research in summarization and identifies missing types of datasets which can be addressed in follow up work.

\subsection{Language and Language Modality}

In the context of summarization, language choice entails the language used for input and output texts. We find a diverse set of datasets in 104 languages (see Appendix~\ref{sec:appendix_langs}).

In addition, we categorize each dataset into three categories of language modality:
\paragraph{Monolingual.} These datasets contain texts in a single language for both input and output.

\paragraph{Multi-lingual.} These datasets can be thought of as a few monolingual datasets joined together, as they contain document-summary pairs in various languages.

\paragraph{Cross-lingual.} In these datasets, the language of the output summary is different from the language of the input document.

\subsection{Domain}

We identified nine domains that encompass the vast majority of datasets: \emph{News}, which relies on articles from online platforms; \emph{Dialogue}, which includes sources such as meeting transcripts, movie scenes, interviews, and customer service interactions; \emph{Scientific} papers, most commonly paired with their abstract as a summary; \emph{Legal} documents such as government bills and patent documents; \emph{Encyclopedia} entries, primarily based on Wikipedia; \emph{Opinions and arguments} including reviews, editorials, and debates; \emph{Social media} texts, which are typically more freeform; \emph{Literature} such as books and poetry; and \emph{Instructional} content, like "how-to" guides.

\begin{table*}[tb!]
\resizebox{\textwidth}{!}{%
\small

\begin{tabular}{@{}p{2.5cm}p{13cm}@{}}
\toprule
\textbf{Shape} & \textbf{Example}                                                                             \\ \midrule
Paragraph(s) &
  Canadian authorities say she was being sought for extradition to the US, where the company is being investigated for possible violation of sanctions against Iran. Canada’s justice department said Meng was arrested in Vancouver on Dec. 1... 
  \\
  \midrule
One Sentence  & John Higgins believes he has the ability and the form to win his fifth World Championship.    \\ 
  \midrule
Span &
  \textbf{Earlier this year the art world made a rare discovery – a painting by Leonardo da Vinci}. Only some 15 paintings by Leonardo still exist, including the “Mona Lisa”. \textbf{So, this painting – known as “Salvator Mundi” or “Savior of the World” – is a truly remarkable find.}  \\ \midrule

Highlights &
    - Web site operator wants to publish encyclopedia about Harry Potter novels. \newline 
     - Judge awards "Harry Potter" author J.K. Rowling and publisher \$6,750 in damages.
     \newline
    - Rowling says she has long planned to publish her own encyclopedia.\\
 \bottomrule
\end{tabular}%
}
\caption{The different shapes of summaries in our collection. Examples are taken from \cite{cheng2016neural,hermann2015teaching}, \cite{lins2019cnn}, \cite{narayan2018don} and \cite{fabbri-etal-2019-multi}. }
\label{tab:types}

\end{table*}

\subsection{Summary Shape}

We categorize the shape of the summaries into four broad types, exemplified in Table~\ref{tab:types}.

\paragraph{Single-sentence summary.}
These are summaries ranging from a few words (like titles) to a full sentence, composed of several dozens of words. This shape is sometimes referred to as ``extreme summarization''~\citep{narayan2018don} or ``headline generation''~\citep{napoles2012annotated}. 

\paragraph{Highlights summary.}
These summaries  are constructed from a few, possibly independent sentences. For example, they could be bullet points from news articles. 

\paragraph{Paragraph summary.}

These datasets contain output texts composed of one or more paragraphs of coherent free text. Examples of this shape include abstracts of scientific papers and sub-headlines of news articles.

\paragraph{Span summary.} Built to accommodate extractive summarization, these datasets use spans of text directly taken from the document as summaries.

\subsection{Sources for Supervision and Annotation Efforts}

We categorize datasets by how much manual annotation effort they require. On opposite ends are datasets that were gathered in an \textbf{Automatic} way, mainly through scraping web resources, and \textbf{Human Annotated} datasets --- in which annotators wrote the gold summaries. We also find \textbf{Semi-Automatic} datasets --- in which human annotators are involved in the process in varying degrees. Datasets created using this method typically provide annotators with summary options that they can correct, filter, or accept.

An additional often overlooked dimension is the objective for which the summaries were originally created, marked as \emph{Supervision} in Figure~\ref{fig:onto}, and consisting of the following possible values:

\paragraph{Natural.} These datasets utilize summaries that were organically created. For example, highlights in news articles  ~\cite{cheng2016neural,hermann2015teaching}  
and educational summaries published to help students~\cite{ladhak2020exploring}. 

\paragraph{Distant.} Similarly to the Natural category, these datasets use pre-existing texts as annotations. However, these texts were \textit{not} originally created to serve as summaries, such as abstracts of scientific papers~\cite{cohan2018discourse} or titles and sub-headlines of news articles~\cite{napoles2012annotated}.

\paragraph{Dedicated.} This category features datasets in which the summaries were written especially for the dataset. It includes mostly human annotated datasets, but also synthetically created ones.

\subsection{Availability}
The distribution of datasets denotes how accessible they are to the research community. We identify three main forms of data availability. 

\paragraph{Publicly Available.} These datasets are often published through a platform like Hugging Face. 

\paragraph{URL-based reconstruction.} In this category, the data is not directly published; instead, a method to reconstruct it is provided. For example, a list of URLs as done in \cite{grusky2018newsroom}, or data preparation scripts, as in \cite{kryscinski2021booksum}. 

\paragraph{Upon request.} This availability type, which is most restricting, usually requires filling out a request form or contacting the authors in order to receive the data.

\section{The State of Summarization Datasets}
\label{Findings}

\begin{figure*}
    \centering 
    \includegraphics[width=\textwidth]{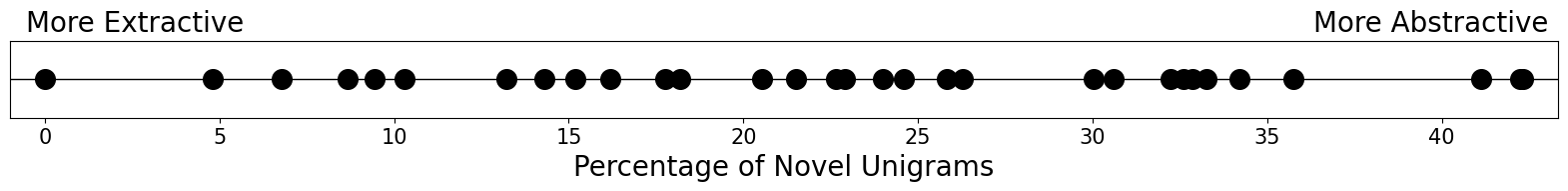}  
    \caption{Every dot on the graph corresponds to a dataset in our collection. Each dot's position indicates the percentage of unique uni-grams in the summaries that are not found in the source texts. Only a minority of the datasets report an evaluation of the properties of the summaries.} 
    \label{fig:spectrum}
\end{figure*}

We thoroughly examine 133 summarization datasets under the lens of our ontology. Figure~\ref{fig:onto} presents the quantitative dataset distribution as well as categorization examples for several datasets.
Following, we discuss five major observations regarding the current state and challenges in summarization datasets.

\subsection{Lack in Common Terminology} 
While summarization is a versatile and expressive task applicable in many scenarios, we find that summarization datasets are presented using inconsistent terminology and often lacking details. This makes it difficult for researchers to identify available resources and to effectively compare them.

For example, determining the \emph{shape} of the summaries (such as paragraph or highlights) often requires accessing the dataset and examining samples; understanding the  \emph{supervision}, how the summaries were created, frequently involved visiting the original website from which the data was scraped; while datasets often do not mention the language in which they were annotated, in contrast with recent emphasis on explicit language disclosure~\cite{bender2019rule}. In our collection, 39 papers failed to specify their languages, often assuming that English is the default language.

All of these highlight the necessity of a shared ontology expressive enough to capture important aspects in summarization datasets, and guide dataset developers in reporting their work.

\paragraph{Abstractive vs. extractive summarization is no longer a useful distinction.}

Another example where terminology is lacking is the distinction between \emph{extractive} summarization, where models are required to select text from the source document and \emph{abstractive} summarization, where generative models can produce novel text.

With the advent of powerful generative LLMs, our survey indicates that these distinctions are no longer useful, with many of the surveyed datasets not specifying if their annotations were extractive or abstractive.

Instead, we argue that \emph{abstractiveness} is a spectrum rather than a binary trait.
To support this, in Figure~\ref{fig:spectrum} we plot the novel n-grams ratio for datasets in our collection. 
This metric, which we find to be the most commonly used for measuring levels of abstractions, measures the percentage of n-grams appearing in the summaries but not in the document \cite{narayan2018don}. Formally:
\begin{equation}
\frac{\text{\# New n-grams}}{\text{\# n-grams}}
\end{equation}
On this scale, 0\% can be achieved  by restricting the summary to contain only vocabulary from the source document and thus lean towards extractive. While at the other end of the scale, 40\% of the 1-grams in the summaries were novel and thus leaning towards abstractive. Other metrics for measuring abstractiveness include Coverage and Density \cite{grusky2018newsroom} as well as Abstractivity (ABS; \citealp{bommasani2020intrinsic}). All of which are based on the overlap between the summary and the source document.


\subsection{News is the Most Popular Domain, Despite its Major Shortcomings}
Roughly 50\% of the datasets in our survey use news articles as source documents. 
This popularity may be due to the large number of online news platforms in various languages that allow for easy scraping. Moreover, many of these articles already include human-written summaries or distant supervision, in the form of highlights, titles, and sub-headlines. 
This wide availability and relative ease of obtaining distant supervision is in fact a \emph{double-edged sword}, harming the quality of the gold summaries, while presenting a rather simple challenge for summarization models.

First, the news domain may present a simple instance of the summarization task. News articles typically follow an
``inverted pyramid writing style'' --- i.e., the most important information answering the 5 w's (who? what? etc.) appears in the first few sentences~\citep{po2003news}. Consequently, a summarization model that trivially outputs the first sentences of a news article can often achieve good results~\citep{cohen2021wikisum}.

Second, news summarization datasets are prone to be of low quality, because they often rely on automatic heuristics. \citet{tejaswin2021well} manually evaluated two datasets with one sentence summaries,
one that uses the title as a summary \cite{rush-etal-2015-neural}, and another that uses a bold sentence at the beginning of the article \cite{narayan2018don}. They found that about half of the samples are either incomplete or contain concepts or entities that do not appear in the rest of the document, and hence often cannot be deduced from it. 

Beyond the news domain, the scientific domain, characterized by longer texts and complex terminology is popular~\cite{cohan2018discourse}, yet it remains predominantly English-centric. Official documents, such as those in the legal domain, patents, and government bills (e.g.,~\citealp{sharma2019bigpatent,kornilova2019billsum,huang2021efficient}), share similar complexity in that salient information is usually distributed throughout the document, and are more multilingual. We also find that dialogues are increasingly attracting the attention of the research community, with this domain being the second most popular, and that Wikipedia articles are a valuable resource due to their guidelines and API, and they are useful for cross-lingual research as they cover the same topics in multiple languages. 

\begin{figure}[tb!]
    \centering
    \includegraphics[width=\columnwidth]
{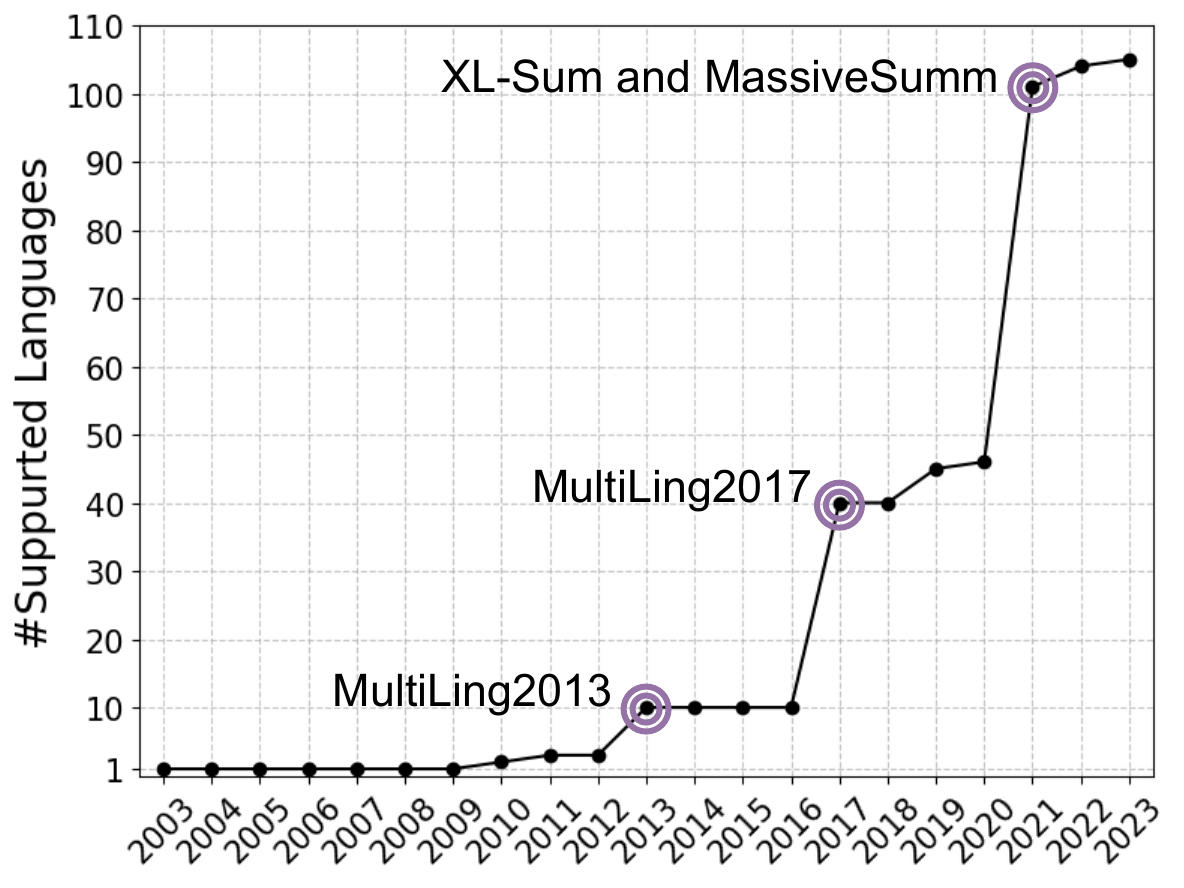}
\caption{The rise in the number of languages supported by summarization resources was achieved mainly through multilingual datasets. Most languages in our survey are only supported through them.}
    \label{fig:mult}
\end{figure}

\subsection{Datasets for Low-Resource Languages are Often Low Quality and Hard to Discover} 
The attention to multilingual NLP has been increasing in recent years~\citep{joshi2020state,ruder2022statemultilingualai}.
Evidently, while English is by far the most popular language in our survey, we cover resources for 104 different languages (see Appendix~\ref{sec:appendix_langs} for the complete list). We find that this encouraging versatility was achieved mainly through multi-lingual datasets, as 81 of the languages surveyed are supported solely through them. In Figure~\ref{fig:mult}, we illustrate the growth in the number of languages with available resources over the years, emphasizing the dramatic rise following the introduction of multilingual resources.

However, we identify limitations in how multilingual datasets are created and reported.

\paragraph{Discovering datasets in low-resource languages proves challenging.}
Given that the majority of languages in our survey are available only within large multilingual corpora, researchers must sift through all multilingual datasets to find their language. This is challenging as there is currently no unified platform that houses all available datasets and allows searching by the desired language.

\begin{figure}[b!]
    \centering 
    \includegraphics[width=\columnwidth]{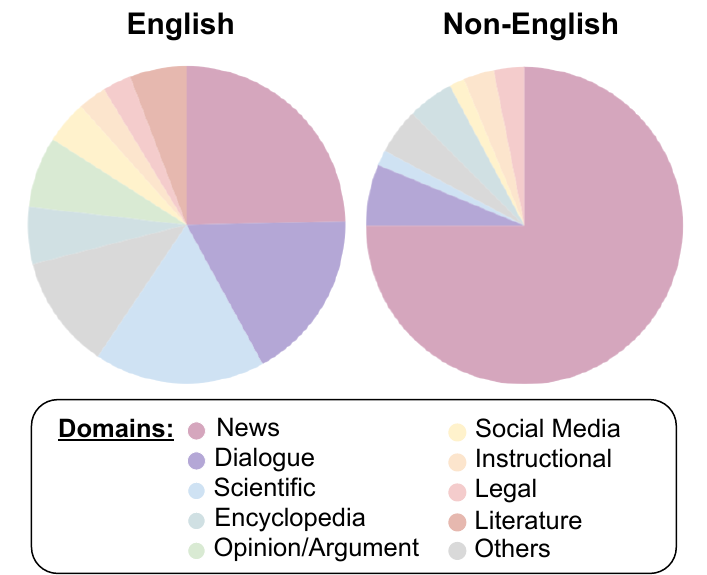}  
    \caption{The distribution of domains for English datasets versus non-English datasets. While English domains are diverse, our survey shows most datasets for other languages are comprised of news articles. }
    \label{fig:domains}

\end{figure}


\paragraph{Datasets in low-resource languages are less diverse in terms of domain, and of poorer quality.}
We find that the over-reliance on the news domain is considerably worse in non-English datasets. In Figure~\ref{fig:domains}, we compare the domain distribution between English and non-English resources. We find that more than 70\% of datasets for these languages are from the news domain. 
This exacerbates the problems we observed regarding the news domain. Indeed, \citet{urlana2022tesum} reported that 97\% of the Telugu samples in XL-SUM~\cite{hasan2021xl} and MassiveSum~\cite{varab2021massivesumm} were of low quality, with similar results in Hindi, Gujarati and Marathi.

\subsection{Annotations Were Often Not Intended to Be Used as Summaries}

We categorize 43\% of the datasets as \emph{Distant} supervision, meaning human authors did not explicitly intend to write a summary of the source document. This is the most common \emph{supervision} category. It includes proxies of summaries like titles and sub-headlines of news articles, abstracts of scientific papers or descriptions of videos and interviews. Other approaches we consider as distant are social media metadata as summaries, as introduced in Newsroom~\cite{grusky2018newsroom} and starting with summaries before searching for longer texts to support them~\cite{ghalandari2020large, saxena2024select}.

On the other end of the supervision spectrum are \emph{dedicated} datasets, where the annotations were created specifically for the dataset and did not exist before. We found only 35 dedicated datasets, three times fewer than datasets that include existing annotations. 

In addition to the vast majority of annotations being gathered automatically, datasets collected this way are also an order of magnitude bigger: human-annotated datasets are on average more than 10 times smaller in number of samples than automatically collected datasets and thus allow for training big models. 

\begin{figure}
    \centering
    \includegraphics[width=\columnwidth]{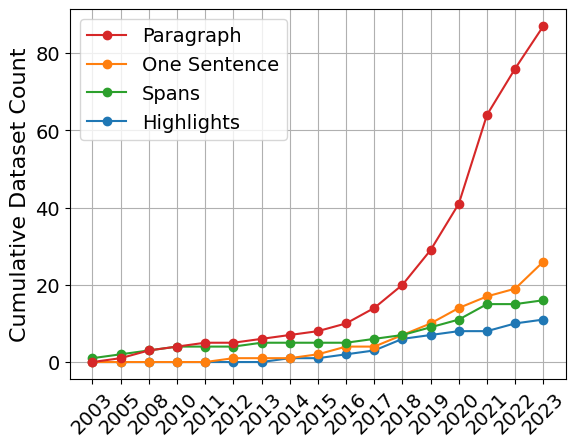}
    \caption{There is a clear trend of collecting summaries of paragraph shape. We note some datasets offer a few shapes.} 
    \label{fig:shape}
\end{figure}

We also find a mismatch between the datasets created and those used. In Figure~\ref{fig:shape}, we show a clear trend over the years of collecting the \emph{paragraph} shape. This might be considered a natural shape for summarization as it is also the most common for collecting data through human annotation in our collection. This comes in contrast to Highlights being very commonly used in research. This is evident from CNN-Daily Mail, perhaps the most popular summarization dataset at the time of writing, which uses this shape.




\subsection{Copyright Concerns Limit the Distribution of Datasets} 
Our survey shows that there is a recent trend moving away from publishing datasets in a public manner, instead making them available through reconstruction from URLs~\citep{grusky2018newsroom,varab2020danewsroom,kryscinski2021booksum}, which may be due to growing legal concerns~\citep{karamolegkou-etal-2023-copyright}.

Reconstruction may be a temporary solution, which does not address the underlying legal issues. Moreover, it may lead to inconsistencies between local instances of the same dataset obtained at different times, due to changes in URL contents. This issue is sometimes mitigated by using snapshots.

Even datasets which were previously publicly available are being called into question.
For example, the popular CNN/Daily Mail and XSUM are considered to be ``legally questionable''~\cite{wang2022squality}. 
In a similar vein, the New York Times Corpus~\cite{sandhaus2008new}, despite being published by the newspaper itself, was recently withdrawn and is no longer available. This may be another downside of the field's over-reliance on the news domain.


\label{04_data_properties}

\section{Standardizing Summarization Datasets} 
\label{sec:fate}

Our survey surfaces a lack in terminology and reporting across the vast number of summarization datasets. This challenges researchers both in determining whether there exists a dataset specifically tailored to their needs (e.g., in terms of language, domain, or shape), and in identifying bottlenecks and directions for future research. Here we outline two contributions we make towards making datasets more standard and accessible. 
\begin{figure}[b!]
    \includegraphics[width =\columnwidth]{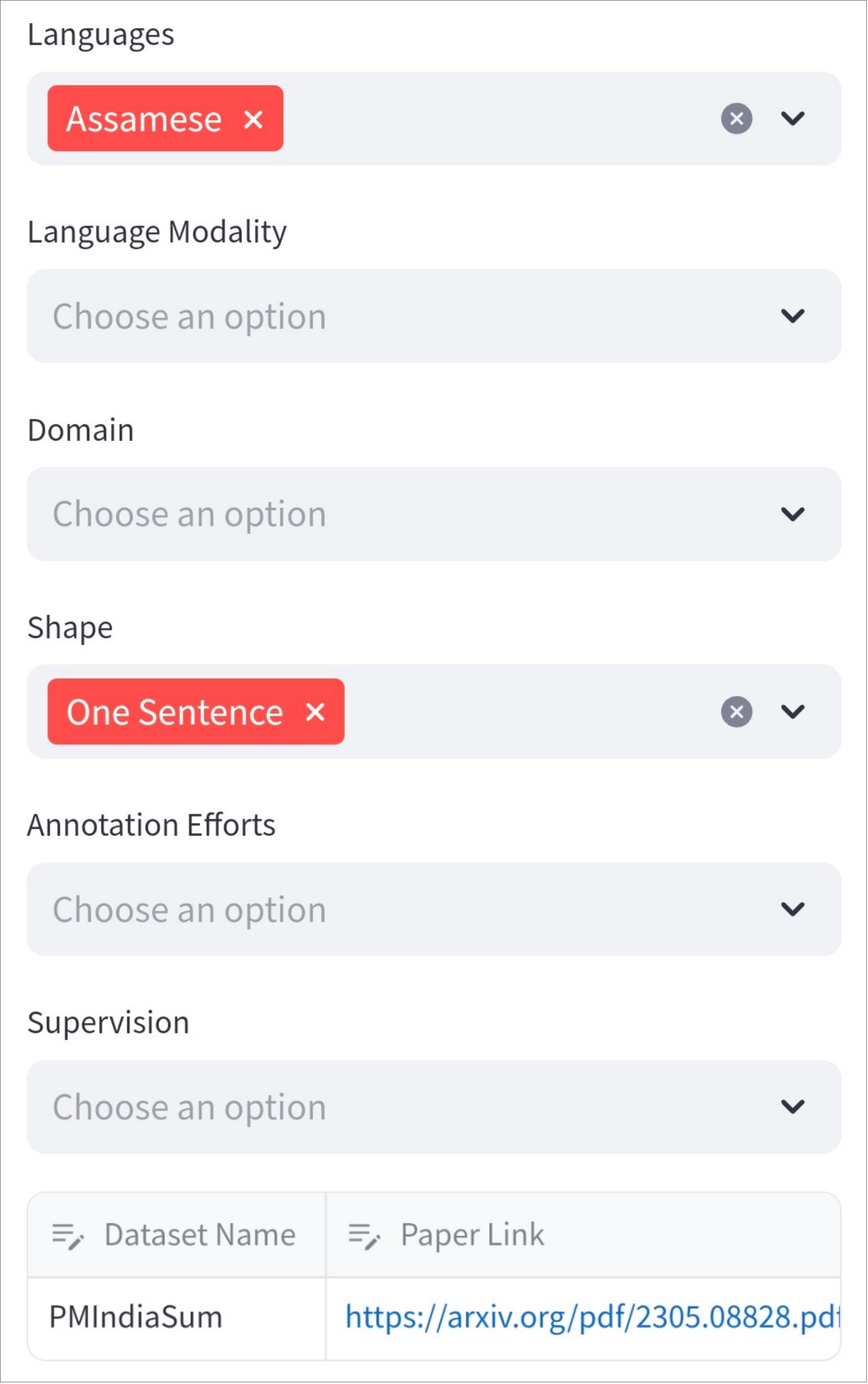}
    \caption{Our interface allows researchers to determine if their specific needs are met by available resources. It is especially important for low resource languages.
    }
    \label{fig:screenshot}
\end{figure}

\paragraph{Interactive web interface.}
We make available a web UI populated with our ontology of 133 summarization datasets,  along with our annotations of the seven axes that characterize each dataset (see Figure~\ref{fig:screenshot}). Users of this interface can search for datasets by specifying the criteria they are looking for. We welcome community contributions to add new datasets, or past datasets not already included in it. We hope that this interface can allow users to easily find relevant datasets, without the need to manually sift through hundreds of papers.

\paragraph{Template for a summarization \emph{data card}.}
In recent years, model and data cards have helped set agreed reporting standards, formalizing what details are reported when publishing a new resource~\citep{model-cards}.
Moving forward, we suggest that summarization research adopt a similar approach. In Table~\ref{tab:datacard} we propose a template for a summarization data card which follows findings from our survey.

Specifically, we focus on four key elements to report: (1) \emph{sample information}: shape, languages, domain; (2) \emph{annotation information}, indicating how were the summaries created: e.g., via scraping or manual effort; (3) \emph{data quality assessment}: we  propose to report the level of abstraction via n-gram novelty, the Compression rate~\citep{grusky2018newsroom}, as well as human evaluation of the data, where applicable;  and (4) \emph{availability}: indicating copyright licensing and download information.

\section{Related Work}

\begin{table}[tb!]
\resizebox{\columnwidth}{!}{%
\begin{tabular}{|p{7.5cm}|}

\hline
\textbf{Summarization Data Card}                                                                                      \\ \hline
\textbf{\underline{Sample information:}}      
\\
\textbf{Languages:} 
\newline
\textit{List all supported languages}                                                                               \\
\textbf{Summary Shape:}
\newline
\textit{Paragraph/One Sentence/Highlights/Span}                                                                \\
\textbf{Domain:} 
\newline
\textit{Example: News/Scientific/Dialogues/etc.}                                                                       \\
\textbf{Size:}
\newline
\textit{Number of document-summary pairs}                                                                                \\ \hline
\textbf{\underline{Annotation information:}}                                                                                      \\
\begin{tabular}[|p{7.5cm}|]{@{}l@{}}\textbf{Annotation efforts:} \\ \textit{Automatic, Human annotations, Semi-automatic}\end{tabular}   \\
\begin{tabular}[|p{7.5cm}|]{@{}l@{}}\textbf{Source of supervision:}\\ \textit{Natural} (summaries created organically)/ \\ \textit{Distant} (annotations are proxies of summaries)/\\ \textit{Dedicated} (annotations created by researchers)\end{tabular} \\
\begin{tabular}[|p{7.5cm}|]{@{}l@{}}\textbf{Brief description of the summaries' source:} \\ \textit{Example:
digests of legal documents}\end{tabular} \\ \hline
\textbf{\underline{Data quality assessment:}}                                                                                     \\
\begin{tabular}[|p{7.5cm}|]{@{}l@{}}\textbf{Abstraction level:} \\ \textit{1-to-4-gram ratios} \end{tabular} \\
\textbf{Compression rate:}  
$ \frac{\text{doc length (\# words)}}{\text{summary length (\# words)}}$                                                                                                    \\
\textbf{Human evaluation:} \textit{Yes/No}                                                                                              \\ \hline
\textbf{\underline{Availability details:}}                                                                                        \\
\begin{tabular}[|p{7.5cm}|]{@{}l@{}}\textbf{How is the data made accessible:} \\ \textit{Publicly Available} / \\
\textit{URL-based Reconstruction} / \\ \textit{Upon Request}\end{tabular}    \\
\begin{tabular}[c]{@{}l@{}}\textbf{Copyrights information:} \\ \textit{License}\end{tabular}                                            \\ \hline
\end{tabular}%
}
\caption{Template for a summarization data card.}
\label{tab:datacard}

\end{table}

Several surveys have explored different aspects of automatic summarization \cite{kryscinski2019neural,cajueiro2023comprehensive,sharma2022automatic}, yet they focus on models rather than datasets. To the best of our knowledge, none of the previous work has focused on the state of low-resource languages, nor looked at the dataset collection method and distribution. 
\citet{dernoncourt2018repository} published a repository of available corpora for summarization. However, it included only a limited number of datasets (21), with just four that are not in English.

Several surveys focus on a sub-task of summarization: \citet{koh2022empirical} reviewed long document summarization; \citet{jangra2023survey} studied multimodal summarization; and \citet{tuggener2021we} conducted a survey of dialogue summarization datasets.

A few papers focused on assessing the properties of popular summarization datasets. \citet{bommasani2020intrinsic} conducted an evaluation of 10 famous datasets using novel metrics and \citet{tejaswin2021well} manually analysed 600 samples from three datasets.

\section{Conclusion}

We carried out an extensive survey of 133 summarization datasets in 104 languages, introducing a new ontology covering sample information, collection methods and data distribution. We offer a detailed report on the current state of resources in the field, surfacing challenges due to the lack of terminology, as well as findings on low resource languages, data quality and availability. To support researchers, we release two resources: a platform to search available datasets by our categorization and a template for a summarization data card, to help streamline future datasets.

\section{Limitations}
Although this survey provides a thorough overview of the available summarization datasets, we wish to acknowledge several limitations. 
First, despite our best efforts, we may not have found all available resources. This is especially true for low resource languages and datasets published by less-established conferences.

Second, we decided to omit from our ontology some axes. Specifically, we did not analyse data about sub-tasks of summarizations such as multi-modal, long-form or aspect based. We note that the available resources are still included in our platform and marked accordingly.

\section*{Acknowledgments}
We would like to thank Omer Kidron, Daria Lioubashevski, the members of the SLAB group at the Hebrew University for the productive discussion, and the anonymous reviewers for their valuable feedback. This works was partially supported by research grants no. 2088, 2336, both from the Israeli Ministry of Science and
Technology.

\bibliography{acl_latex}

\appendix

\section{Datasets Categorized by Language}
\label{sec:appendix_langs}
In this section we list all datasets included in this survey per language. 

\paragraph{Afrikaans:}
\cite{varab2021massivesumm}, \cite{giannakopoulos2017multiling}
\paragraph{Albanian:}
\cite{varab2021massivesumm}, \cite{palen2212lr}
\paragraph{Amharic:}
\cite{hasan2021xl}, \cite{palen2212lr}, \cite{bhattacharjee2021crosssum}, \cite{shi2023mcls}, \cite{varab2021massivesumm}, \cite{liang2022summary}, \cite{bhattacharjee2021crosssum}, \cite{xiaorui2023mcls}
\paragraph{Arabic:} \cite{giannakopoulos2015multiling}, \cite{el2010using}, \cite{hasan2021xl}, \cite{nguyen2019global}, \cite{el2013kalimat}, \cite{alhamadani2022lans}, \cite{chouigui2021arabic}, \cite{varab2021massivesumm}, \cite{liang2022summary}, \cite{giannakopoulos2017multiling}, \cite{ladhak2020wikilingua}, \cite{giannakopoulos2017multiling}, \cite{bhattacharjee2021crosssum}, \cite{tikhonov2022wikimulti}, \cite{xiaorui2023mcls}
\paragraph{Armenian:} \cite{palen2212lr}, \cite{varab2021massivesumm}
\paragraph{Assamese:} \cite{urlana2023pmindiasum}, \cite{goutom2023abstractive}, \cite{varab2021massivesumm}
\paragraph{Aymara:} \cite{varab2021massivesumm}
\paragraph{Azerbaijani:} \cite{hasan2021xl}, \cite{palen2212lr}, \cite{varab2021massivesumm}, \cite{liang2022summary}, \cite{giannakopoulos2017multiling},\cite{bhattacharjee2021crosssum}, \cite{xiaorui2023mcls}
\paragraph{Bambara:} \cite{varab2021massivesumm}
\paragraph{Basque:} \cite{giannakopoulos2017multiling}
\paragraph{Bengali:} \cite{hasan2021xl}, \cite{urlana2023pmindiasum}, \cite{palen2212lr}, \cite{nguyen2019global}, \cite{khan2023banglachq}, \cite{varab2021massivesumm}, \cite{liang2022summary}, \cite{verma2023large}, \cite{bhattacharjee2021crosssum}, \cite{xiaorui2023mcls}
\paragraph{Bislama:} \cite{varab2021massivesumm}
\paragraph{Bosnian:} \cite{palen2212lr}, \cite{varab2021massivesumm}, \cite{cao2020multisumm}, \cite{giannakopoulos2017multiling}
\paragraph{Bulgarian:} \cite{varab2021massivesumm}, \cite{giannakopoulos2017multiling}, \cite{aumiller2022eur}
\paragraph{Burmese:} \cite{hasan2021xl}, \cite{palen2212lr}, \cite{varab2021massivesumm}, \cite{liang2022summary}, \cite{bhattacharjee2021crosssum}, \cite{xiaorui2023mcls}
\paragraph{Catalan:} \cite{soriano2022dacsa}, \cite{varab2021massivesumm}, \cite{giannakopoulos2017multiling}
\paragraph{Central Khmer:} \cite{palen2212lr}, \cite{varab2021massivesumm}
\paragraph{Chinese:}\footnote{This list includes both Traditional Chinese and Simplified Chinese} \cite{giannakopoulos2015multiling}, \cite{hu2015lcsts}, \cite{hasan2021xl}, \cite{palen2212lr}, \cite{wu2023vcsum}, \cite{liu2020clts}, \cite{liu2022clts+}, \cite{zang2022newsfarm}, \cite{chen2021large}, \cite{zhao2021qbsum}, \cite{varab2021massivesumm}, \cite{li2017multi}, \cite{liang2022summary}, \cite{lin2021csds}, \cite{verma2023large}, \cite{li2020vmsmo}, \cite{wang2021contrastive}, \cite{giannakopoulos2017multiling}, \cite{ladhak2020wikilingua}, \cite{zhu2019ncls}, \cite{wang2022clidsum}, \cite{bhattacharjee2021crosssum}, \cite{takeshita2024cross}, \cite{tikhonov2022wikimulti}, \cite{xiaorui2023mcls}
\paragraph{Croatian:} \cite{varab2021massivesumm}, \cite{cao2020multisumm}, \cite{giannakopoulos2017multiling}, \cite{aumiller2022eur}
\paragraph{Czech:} \cite{giannakopoulos2015multiling},\cite{straka2018sumeczech}, \cite{varab2021massivesumm}, \cite{giannakopoulos2017multiling}, \cite{ladhak2020wikilingua}, \cite{perez2022models}, \cite{aumiller2022eur}

\paragraph{Danish:} \cite{varab2020danewsroom}, \cite{varab2021massivesumm}, \cite{aumiller2022eur}
\paragraph{Dari:} \cite{palen2212lr}, \cite{varab2021massivesumm}
\paragraph{Dutch:} \cite{nguyen2019global}, \cite{varab2021massivesumm}, \cite{giannakopoulos2017multiling}, \cite{ladhak2020wikilingua}, \cite{aumiller2022eur}, \cite{tikhonov2022wikimulti}
\paragraph{English:} \cite{giannakopoulos2015multiling}, \cite{cheng2016neural,hermann2015teaching}, \cite{grusky2018newsroom}, \cite{hasan2021xl}, \cite{fabbri-etal-2019-multi}, \cite{cachola2020tldr}, \cite{takeshita2024aclsum}, \cite{jiang2024ccsum}, \cite{saxena2024select}, \cite{mahbub2023unveiling}, \cite{urlana2023pmindiasum}, \cite{datta2023mildsum}, \cite{klaus2022summarizing}, \cite{palen2212lr}, \cite{cohen2021wikisum}, \cite{koupaee2018wikihow}, \cite{huang2021efficient}, \cite{sharma2019bigpatent}, \cite{napoles2012annotated}, \cite{nguyen2019global}, \cite{kornilova2019billsum}, \cite{hasler2003building}, \cite{lins2019cnn}, \cite{zhang2019email}, \cite{wang2016neural}, \cite{ganesan2010opinosis}, \cite{cohan2018discourse}, \cite{kim2018abstractive}, \cite{Gliwa_2019}, \cite{zhu2021mediasumlargescalemediainterview}, \cite{yasunaga2019scisummnetlargeannotatedcorpus}, \cite{cao2016tgsum}, \cite{wei2018utilizing}, \cite{narayan2018don}, \cite{abacha2019summarization}, \cite{yadav2022chq}, \cite{xu2021miranews}, \cite{zhang2017towards}, \cite{ghalandari2020large}, \cite{sotudeh2021tldr9+}, \cite{hayashi2021wikiasp}, \cite{malykh2020sumtitles}, \cite{ta2023wikides}, \cite{syed2020news}, \cite{liu2023generating}, \cite{kraaij2005ami}, \cite{volske2017tl}, \cite{chen2021dialogsum}, \cite{gupta2021sumpubmed}, \cite{kryscinski2021booksum}, \cite{ladhak2020exploring}, \cite{hayashi2020s}, \cite{antognini2020gamewikisum}, \cite{wang2022squality}, \cite{feigenblat2021tweetsumm}, \cite{lu2020multi}, \cite{chen2021summscreen}, \cite{meng2021bringing}, \cite{khalman2021forumsum}, \cite{roush2020debatesum}, \cite{zhong2021qmsum}, \cite{bhatia2014summarizing}, \cite{ulrich2008publicly}, \cite{varab2021massivesumm}, \cite{li2018multi}, \cite{li2017multi}, \cite{liang2022summary}, \cite{zhu2018msmo}, \cite{chen2018abstractive}, \cite{yadav2021subsume}, \cite{gorinski2015movie}, \cite{verma2023large}, \cite{sanabria2018how2}, \cite{wang2021contrastive}, \cite{zopf2018auto}, \cite{zopf2016next}, \cite{collins2017supervised}, \cite{lev2019talksumm}, \cite{park2023acl}, \cite{giannakopoulos2017multiling}, \cite{ladhak2020wikilingua}, \cite{zhu2019ncls}, \cite{perez2022models}, \cite{wang2022clidsum}, \cite{aumiller2022eur}, \cite{bhattacharjee2021crosssum}, \cite{takeshita2024cross}, \cite{fatima2021novel}, \cite{tikhonov2022wikimulti}, \cite{xiaorui2023mcls}
\paragraph{Esperanto:} \cite{varab2021massivesumm}, \cite{giannakopoulos2017multiling}
\paragraph{Estonian:} \cite{aumiller2022eur}
\paragraph{Filipino:} \cite{varab2021massivesumm}
\paragraph{Finnish:} \cite{giannakopoulos2017multiling}, \cite{aumiller2022eur}
\paragraph{French:} \cite{giannakopoulos2015multiling}, \cite{hasan2021xl}, \cite{palen2212lr}, \cite{scialom2020mlsum}, \cite{nguyen2019global}, \cite{varab2021massivesumm}, \cite{liang2022summary}, \cite{eddine2020barthez}, \cite{verma2023large}, \cite{wang2021contrastive}, \cite{giannakopoulos2017multiling}, \cite{ladhak2020wikilingua}, \cite{perez2022models}, \cite{bhattacharjee2021crosssum}, \cite{tikhonov2022wikimulti}, \cite{xiaorui2023mcls}

\paragraph{Fulah:} \cite{varab2021massivesumm}
\paragraph{Georgian:} \cite{palen2212lr}, \cite{varab2021massivesumm}, \cite{giannakopoulos2017multiling}
\paragraph{German:}  \cite{scialom2020mlsum}, \cite{nguyen2019global}, \cite{varab2021massivesumm}, \cite{kew202320}, \cite{wang2021contrastive}, \cite{zopf2018auto}, \cite{giannakopoulos2017multiling}, \cite{wang2022clidsum}, \cite{aumiller2022eur}, \cite{takeshita2024cross}, \cite{fatima2021novel}, \cite{tikhonov2022wikimulti}
\paragraph{Gujarati:} \cite{hasan2021xl}, \cite{satapara2022fire}, \cite{urlana2023pmindiasum}, \cite{varab2021massivesumm}, \cite{liang2022summary}, \cite{verma2023large}, \cite{ladhak2020wikilingua}, \cite{perez2022models}, \cite{bhattacharjee2021crosssum}, \cite{xiaorui2023mcls}
\paragraph{Haitian:} \cite{palen2212lr}, \cite{varab2021massivesumm}
\paragraph{Hausa:} \cite{hasan2021xl}, \cite{palen2212lr}, \cite{varab2021massivesumm}, \cite{liang2022summary}, \cite{bhattacharjee2021crosssum}, \cite{xiaorui2023mcls}
\paragraph{Hebrew:} \cite{giannakopoulos2015multiling}, \cite{varab2021massivesumm}, \cite{mondshine2024hesum}
\paragraph{Hindi:} \cite{giannakopoulos2015multiling}, \cite{hasan2021xl}, \cite{satapara2022fire}, \cite{urlana2023pmindiasum}, \cite{datta2023mildsum}, \cite{varab2021massivesumm}, \cite{liang2022summary}, \cite{verma2023large}, \cite{wang2021contrastive}, \cite{ladhak2020wikilingua}, \cite{bhattacharjee2021crosssum}, \cite{xiaorui2023mcls}
\paragraph{Hungarian:} \cite{varab2021massivesumm}, \cite{barta2024news}, \cite{aumiller2022eur}
\paragraph{Icelandic:} \cite{varab2021massivesumm}
\paragraph{Igbo:} \cite{hasan2021xl}, \cite{varab2021massivesumm}, \cite{liang2022summary}, \cite{bhattacharjee2021crosssum}, \cite{xiaorui2023mcls}
\paragraph{Indonesian:} \cite{hasan2021xl}, \cite{palen2212lr}, \cite{koto2020liputan6}, \cite{kurniawan2018indosum},  \cite{varab2021massivesumm}, \cite{liang2022summary}, \cite{verma2023large}, \cite{wang2021contrastive}, \cite{giannakopoulos2017multiling}, \cite{ladhak2020wikilingua}, \cite{bhattacharjee2021crosssum}, \cite{xiaorui2023mcls}
\paragraph{Irish:} \cite{varab2021massivesumm}, \cite{aumiller2022eur}
\paragraph{Italian:} \cite{landro2022two}, \cite{nguyen2019global}, \cite{varab2021massivesumm}, \cite{giannakopoulos2017multiling}, \cite{ladhak2020wikilingua}, \cite{aumiller2022eur}, \cite{takeshita2024cross}, \cite{tikhonov2022wikimulti}
\paragraph{Japanese:} \cite{hasan2021xl},\cite{nguyen2019global}, \cite{varab2021massivesumm}, \cite{liang2022summary}, \cite{yamamura2018annotation}, \cite{verma2023large}, \cite{giannakopoulos2017multiling}, \cite{ladhak2020wikilingua}, \cite{bhattacharjee2021crosssum}, \cite{takeshita2024cross}, \cite{tikhonov2022wikimulti}, \cite{xiaorui2023mcls}
\paragraph{Javanese:} \cite{giannakopoulos2017multiling}
\paragraph{Kannada:} \cite{urlana2023pmindiasum}, \cite{varab2021massivesumm}
\paragraph{Kinyarwanda:} \cite{palen2212lr}, \cite{varab2021massivesumm}
\paragraph{Kirghiz:} \cite{hasan2021xl}, \cite{varab2021massivesumm}, \cite{liang2022summary}, \cite{bhattacharjee2021crosssum}, \cite{xiaorui2023mcls}
\paragraph{Korean:} \cite{hasan2021xl}, \cite{palen2212lr}, \cite{liang2022summary}, \cite{giannakopoulos2017multiling}, \cite{ladhak2020wikilingua}, \cite{bhattacharjee2021crosssum}, \cite{xiaorui2023mcls}
\paragraph{Kurdish:} \cite{palen2212lr}, \cite{badawi2023kurdsum}, \cite{varab2021massivesumm}
\paragraph{Lao:} \cite{palen2212lr}, \cite{varab2021massivesumm}
\paragraph{Latvian:} \cite{varab2021massivesumm}, \cite{giannakopoulos2017multiling}, \cite{aumiller2022eur}
\paragraph{Limburgish:} \cite{giannakopoulos2017multiling}
\paragraph{Lingala:} \cite{varab2021massivesumm}
\paragraph{Lithuanian:} \cite{varab2021massivesumm}, \cite{aumiller2022eur}
\paragraph{Macedonian:} \cite{palen2212lr}, \cite{nguyen2019global}, \cite{varab2021massivesumm}
\paragraph{Malagasy:} \cite{nguyen2019global}, \cite{varab2021massivesumm}
\paragraph{Malay:} \cite{urlana2023pmindiasum}, \cite{giannakopoulos2017multiling}

\paragraph{Malayalam:} \cite{urlana2023pmindiasum}, \cite{varab2021massivesumm}
\paragraph{Maltese:} \cite{aumiller2022eur}
\paragraph{Manipuri:} \cite{urlana2023pmindiasum}
\paragraph{Marathi:} \cite{hasan2021xl}, \cite{urlana2023pmindiasum}, \cite{urlana2023pmindiasum}, \cite{varab2021massivesumm}, \cite{liang2022summary}, \cite{verma2023large}, \cite{giannakopoulos2017multiling}, \cite{bhattacharjee2021crosssum}, \cite{xiaorui2023mcls}
\paragraph{Modern Greek:} \cite{giannakopoulos2015multiling}, \cite{palen2212lr}, \cite{nguyen2019global}, \cite{varab2021massivesumm}, \cite{giannakopoulos2017multiling}, \cite{aumiller2022eur}
\paragraph{Mongolian:} \cite{varab2021massivesumm}
\paragraph{Nepali:} \cite{hasan2021xl}, \cite{varab2021massivesumm}, \cite{liang2022summary}, \cite{verma2023large}, \cite{bhattacharjee2021crosssum}, \cite{xiaorui2023mcls}
\paragraph{North Ndebele:} \cite{palen2212lr}, \cite{varab2021massivesumm}
\paragraph{Norwegian:} \cite{giannakopoulos2017multiling}
\paragraph{Oriya:} \cite{urlana2023pmindiasum}, \cite{varab2021massivesumm}
\paragraph{Oromo:} \cite{hasan2021xl}, \cite{varab2021massivesumm}, \cite{liang2022summary}, \cite{bhattacharjee2021crosssum}, \cite{xiaorui2023mcls}
\paragraph{Panjabi:} \cite{hasan2021xl}, \cite{urlana2023pmindiasum}, \cite{varab2021massivesumm}, \cite{liang2022summary}, \cite{verma2023large}, \cite{bhattacharjee2021crosssum}, \cite{xiaorui2023mcls}

\paragraph{Pashto:} \cite{hasan2021xl}, \cite{palen2212lr}, \cite{varab2021massivesumm}, \cite{liang2022summary}, \cite{verma2023large}, \cite{bhattacharjee2021crosssum}, \cite{xiaorui2023mcls}
\paragraph{Persian:} \cite{hasan2021xl}, \cite{palen2212lr}, \cite{liang2022summary}, \cite{farahani2021leveraging}, \cite{giannakopoulos2017multiling}, \cite{bhattacharjee2021crosssum}, \cite{xiaorui2023mcls}
\paragraph{Polish:} \cite{varab2021massivesumm}, \cite{giannakopoulos2017multiling}, \cite{aumiller2022eur}, \cite{tikhonov2022wikimulti}
\paragraph{Portuguese:} \cite{hasan2021xl}, \cite{palen2212lr}, \cite{nguyen2019global}, \cite{cardoso2011cstnews}, \cite{paiola2024recognasumm}, \cite{varab2021massivesumm}, \cite{liang2022summary}, \cite{verma2023large}, \cite{sanabria2018how2}, \cite{wang2021contrastive}, \cite{giannakopoulos2017multiling}, \cite{ladhak2020wikilingua}, \cite{aumiller2022eur}, \cite{bhattacharjee2021crosssum}, \cite{tikhonov2022wikimulti}, \cite{xiaorui2023mcls}
\paragraph{Romanian:} \cite{giannakopoulos2015multiling}, \cite{varab2021massivesumm}, \cite{giannakopoulos2017multiling}, \cite{aumiller2022eur}
\paragraph{Rundi:} \cite{hasan2021xl}, \cite{varab2021massivesumm}, \cite{liang2022summary}, \cite{bhattacharjee2021crosssum}, \cite{xiaorui2023mcls}
\paragraph{Russian:} \cite{hasan2021xl}, \cite{palen2212lr},  \cite{scialom2020mlsum}, \cite{nguyen2019global}, \cite{gusev2020dataset}, \cite{varab2021massivesumm}, \cite{liang2022summary}, \cite{verma2023large}, \cite{wang2021contrastive}, \cite{giannakopoulos2017multiling}, \cite{ladhak2020wikilingua}, \cite{bhattacharjee2021crosssum}, \cite{tikhonov2022wikimulti}, \cite{xiaorui2023mcls}
\paragraph{Scottish Gaelic:} \cite{hasan2021xl}, \cite{varab2021massivesumm}, \cite{liang2022summary}, \cite{bhattacharjee2021crosssum}, \cite{xiaorui2023mcls}
\paragraph{Serbian:}\footnote{This list includes Cyrillic and Latin writing systems}  \cite{hasan2021xl}, \cite{palen2212lr}, \cite{varab2021massivesumm}, \cite{liang2022summary}, \cite{bhattacharjee2021crosssum}, \cite{xiaorui2023mcls}
\paragraph{Shona:} \cite{palen2212lr}, \cite{varab2021massivesumm}
\paragraph{Sinhala:} \cite{hasan2021xl}, \cite{varab2021massivesumm}, \cite{liang2022summary}, \cite{verma2023large}, \cite{bhattacharjee2021crosssum}, \cite{xiaorui2023mcls}

\paragraph{Slovak:} \cite{ondrejova2024slovaksum}, \cite{vsuppa2020summarization}, \cite{varab2021massivesumm}, \cite{giannakopoulos2017multiling}, \cite{aumiller2022eur}
\paragraph{Slovenian:} \cite{varab2021massivesumm}, \cite{aumiller2022eur}
\paragraph{Somali:} \cite{hasan2021xl}, \cite{palen2212lr}, \cite{varab2021massivesumm}, \cite{liang2022summary}, \cite{bhattacharjee2021crosssum}, \cite{xiaorui2023mcls}

\paragraph{Spanish:} \cite{giannakopoulos2015multiling}, \cite{hasan2021xl}, \cite{soriano2022dacsa}, \cite{palen2212lr}, \cite{scialom2020mlsum}, \cite{nguyen2019global}, \cite{varab2021massivesumm}, \cite{liang2022summary}, \cite{verma2023large}, \cite{wang2021contrastive}, \cite{giannakopoulos2017multiling}, \cite{ladhak2020wikilingua}, \cite{bhattacharjee2021crosssum}, \cite{tikhonov2022wikimulti}, \cite{xiaorui2023mcls}

\paragraph{Swedish:} \cite{varab2021massivesumm}, \cite{aumiller2022eur}, \cite{tikhonov2022wikimulti}
\paragraph{Tagalog:} \cite{giannakopoulos2017multiling}

\paragraph{Tajik:} \cite{varab2021massivesumm}
\paragraph{Tamil:} \cite{hasan2021xl}, \cite{urlana2023pmindiasum}, \cite{varab2021massivesumm}, \cite{liang2022summary}, \cite{verma2023large}, \cite{bhattacharjee2021crosssum}, \cite{xiaorui2023mcls}

\paragraph{Telugu:} \cite{hasan2021xl}, \cite{urlana2023pmindiasum}, \cite{urlana2022tesum}, \cite{varab2021massivesumm}, \cite{liang2022summary}, \cite{verma2023large}, \cite{bhattacharjee2021crosssum}, \cite{xiaorui2023mcls}

\paragraph{Tetum:} \cite{varab2021massivesumm}
\paragraph{Thai:} \cite{hasan2021xl}, \cite{palen2212lr}, \cite{varab2021massivesumm}, \cite{liang2022summary}, \cite{giannakopoulos2017multiling}, \cite{ladhak2020wikilingua}, \cite{bhattacharjee2021crosssum}, \cite{xiaorui2023mcls}

\paragraph{Tibetan:} \cite{palen2212lr}, \cite{varab2021massivesumm}
\paragraph{Tigrinya:} \cite{hasan2021xl}, \cite{palen2212lr}, \cite{varab2021massivesumm}, \cite{liang2022summary}, \cite{bhattacharjee2021crosssum}, \cite{xiaorui2023mcls}

\paragraph{Turkish:} \cite{hasan2021xl}, \cite{palen2212lr}, \cite{scialom2020mlsum}, \cite{varab2021massivesumm}, \cite{liang2022summary}, \cite{wang2021contrastive}, \cite{giannakopoulos2017multiling}, \cite{ladhak2020wikilingua}, \cite{bhattacharjee2021crosssum}, \cite{xiaorui2023mcls}

\paragraph{Ukrainian:} \cite{hasan2021xl}, \cite{palen2212lr}, \cite{galeshchuk2023abstractive}, \cite{varab2021massivesumm}, \cite{liang2022summary}, \cite{verma2023large}, \cite{wang2021contrastive}, \cite{giannakopoulos2017multiling}, \cite{bhattacharjee2021crosssum}, \cite{tikhonov2022wikimulti}, \cite{xiaorui2023mcls}

\paragraph{Urdu:} \cite{hasan2021xl}, \cite{urlana2023pmindiasum}, \cite{palen2212lr}, \cite{faheem2024urdumasd}, \cite{varab2021massivesumm}, \cite{liang2022summary}, \cite{verma2023large}, \cite{bhattacharjee2021crosssum}, \cite{xiaorui2023mcls}

\paragraph{Uzbek:} \cite{hasan2021xl}, \cite{palen2212lr}, \cite{varab2021massivesumm}, \cite{liang2022summary}, \cite{bhattacharjee2021crosssum}, \cite{xiaorui2023mcls}

\paragraph{Vietnamese:} \cite{hasan2021xl}, \cite{palen2212lr}, \cite{nguyen2019vnds}, \cite{tran2020vims}, \cite{varab2021massivesumm}, \cite{liang2022summary}, \cite{wang2021contrastive}, \cite{ladhak2020wikilingua}, \cite{bhattacharjee2021crosssum}, \cite{tikhonov2022wikimulti}, \cite{xiaorui2023mcls}

\paragraph{Welsh:} \cite{hasan2021xl}, \cite{varab2021massivesumm}, \cite{liang2022summary}, \cite{bhattacharjee2021crosssum}, \cite{xiaorui2023mcls}

\paragraph{West African Pidgin English:} \cite{hasan2021xl}, \cite{liang2022summary}, \cite{bhattacharjee2021crosssum}, \cite{xiaorui2023mcls}
\paragraph{Xhosa:} \cite{varab2021massivesumm}
\paragraph{Yoruba:} \cite{hasan2021xl}, \cite{varab2021massivesumm}, \cite{liang2022summary}, \cite{bhattacharjee2021crosssum}, \cite{xiaorui2023mcls}

\paragraph{Swahili:} \cite{hasan2021xl}, \cite{palen2212lr}, \cite{nguyen2019global}, \cite{varab2021massivesumm}, \cite{liang2022summary}, \cite{bhattacharjee2021crosssum}, \cite{xiaorui2023mcls}




\end{document}